\documentclass{article}

\usepackage[sglblindworkshop, final]{neurips_2026}

\usepackage[utf8]{inputenc} 
\usepackage[T1]{fontenc}    
\usepackage{hyperref}       
\usepackage{url}            
\usepackage{amsfonts}       
\usepackage{nicefrac}       
\usepackage{microtype}      

\usepackage{wrapfig}
\usepackage{colortbl}
\usepackage{pgfplots}
\pgfplotsset{compat=1.18}

\usepackage{booktabs}
\usepackage{graphicx}
\usepackage[table]{xcolor}
\usepackage{pgf}

\makeatletter
\def\asrcell@delta{delta}
\def\asrcell@asr{asr}
\def\asrcell@refusal{refusal}
\def\asrcell@refusaldelta{refusal-delta}
\NewDocumentCommand{\asrcell}{O{delta} m o}{%
  \def\asrcell@mode{#1}%
  \def\asrcell@fill{white}%
  \ifx\asrcell@mode\asrcell@asr
    \ifdim#2pt<20pt \def\asrcell@fill{green!20}%
    \else\ifdim#2pt<60pt \def\asrcell@fill{orange!15}%
    \else\ifdim#2pt<80pt \def\asrcell@fill{red!15}%
    \else \def\asrcell@fill{red!25}%
    \fi\fi\fi
  \else\ifx\asrcell@mode\asrcell@refusal
    \ifdim#2pt<20pt \def\asrcell@fill{red!25}%
    \else\ifdim#2pt<40pt \def\asrcell@fill{red!15}%
    \else\ifdim#2pt<80pt \def\asrcell@fill{orange!15}%
    \else \def\asrcell@fill{green!20}%
    \fi\fi\fi
  \else\ifx\asrcell@mode\asrcell@refusaldelta
    \ifdim#2pt>0pt \def\asrcell@fill{green!20}%
    \else\ifdim#2pt=0pt \def\asrcell@fill{white}%
    \else\ifdim#2pt>-20pt \def\asrcell@fill{orange!15}%
    \else\ifdim#2pt>-40pt \def\asrcell@fill{red!15}%
    \else \def\asrcell@fill{red!25}%
    \fi\fi\fi\fi
  \else
    \ifdim#2pt<0pt \def\asrcell@fill{green!20}%
    \else\ifdim#2pt=0pt \def\asrcell@fill{white}%
    \else\ifdim#2pt<20pt \def\asrcell@fill{orange!15}%
    \else\ifdim#2pt<40pt \def\asrcell@fill{red!15}%
    \else \def\asrcell@fill{red!25}%
    \fi\fi\fi\fi
  \fi\fi\fi
  \expandafter\cellcolor\expandafter{\asrcell@fill}%
  $#2\IfValueT{#3}{\,\pm\,#3}$%
}
\makeatother

\usepackage{bm}
\usepackage{tcolorbox}
\tcbuselibrary{skins}

\definecolor{sand}{RGB}{248,245,235}  
\definecolor{sandborder}{RGB}{220,215,200}
\definecolor{darkblue}{RGB}{40,70,140}   
\definecolor{parchment}{RGB}{252,249,240}
\definecolor{parchmentborder}{RGB}{222,215,195}
\definecolor{sage}{RGB}{244,248,244}
\definecolor{sageborder}{RGB}{200,210,200}
\definecolor{lavender}{RGB}{247,246,250}
\definecolor{lavenderborder}{RGB}{210,205,225}
\definecolor{riskbg}{RGB}{255,246,246}
\definecolor{riskborder}{RGB}{205,105,105}

\usepackage{enumitem}

\title{Do AI-for-Science Agents Preserve Model Safety?}

\workshoptitle{Third Workshop on Agents in the Wild: Safety, Security, and Beyond}

\author{%
  Michael Yew Chuan Lee \\
  DSO National Laboratories\\
  \And
  Manya Sriram \\
  UC Berkeley \\
  \And
   N. Benjamin Erichson \\
   ICSI and Berkeley Lab \\
}

\begin{document}

\maketitle

\begin{abstract}
AI-for-Science agents are increasingly being used for formulating scientific hypotheses, running autonomous experiments, and interpreting evidence. These agents are constructed by integrating a large language model (LLM) into a domain-specific agent harness, which in turn can improve the LLM's scientific capability and enable the model to orchestrate scientific tools. In this work, we ask whether model safety is preserved when an LLM is integrated in an AI-for-Science agent harness. We evaluate five agent harnesses on 50 harmful chemistry requests. Specifically, we study how the safety behavior changes under direct prompting, two simple adversarial prompting strategies, and an adaptive attack. Across four recent GPT variants, our results show that the agent harness can increase harmful compliance even under direct prompting. More critically, we find that adaptive attacks that are largely unsuccessful against the base LLM suddenly become highly successful after the agent harness is introduced. Our results show that the safety behavior of the base LLM is not durable once a harness is introduced. Thus, as AI-for-Science agents become more capable and autonomous, the community needs to evaluate and align the agent using domain-specific safety benchmarks.
\end{abstract}

\section{Introduction}

AI-for-Science agents based on large language models (LLMs) are becoming an increasingly important part of scientific discovery~\cite{gao2024empowering}. Rather than using an LLM only to answer questions, these systems place the model inside larger workflows that connect it to scientific databases, specialized software, and computational tools~\cite{boiko2023autonomous}. The broader goal is to move toward increasingly autonomous scientific workflows that can support the research process from hypothesis generation to analysis and experimental validation. In chemistry and biology, agents already extend LLMs beyond text generation to tasks such as data analysis, synthesis planning, and multi-step scientific reasoning~\cite{m2024augmenting,huang2025biomni}. These systems can therefore make the same base model more capable in a scientific setting.

For the class of systems considered here, the agent is not a new language model. It is a base LLM embedded in an \emph{agent harness}, the prompting and software layer that orchestrates how the model interacts with its context, tools, and external environment. In many AI-for-Science systems, the model weights are unchanged; what changes is the system around the model. A domain-specific harness can therefore alter the information available to the model, the actions it can take, and the sequence of interactions through which it solves a task. This is how the same base model can become more capable when deployed as a scientific agent. The behavior of the resulting system is therefore a property of the model and harness together, rather than of the model alone.

The agent harness, however, also changes how the model behaves. A request may be presented under different instructions, combined with additional context, revisited over multiple steps, or acted on through tools. These changes matter because refusal behavior is known to be sensitive to how a harmful request is presented and pursued. 
Even simple reformulations can weaken refusals~\cite{andriushchenko2024pasttense}, while adaptive multi-turn interactions can elicit harmful assistance that is not observed in single-turn evaluations~\cite{rahman2025xteaming}. Related work on general-purpose LLM agents has also shown that harmful agentic behavior can emerge in multi-step tasks, including without an explicit jailbreak~\cite{andriushchenko2025agentharm}.
A scientific agent harness therefore does more than increase capability. It can also create new pathways through which harmful requests are interpreted and answered. The refusal behavior observed for the base model may therefore not be preserved once the harness is introduced, even when the model weights themselves are unchanged. This does not imply that every harness makes a model less safe. It means that safety must be evaluated on the composed agent rather than inferred from the base model.

\begin{wrapfigure}[19]{r}{0.47\textwidth}
    \centering
    \vspace{-1.7em}
    \includegraphics[width=\linewidth]{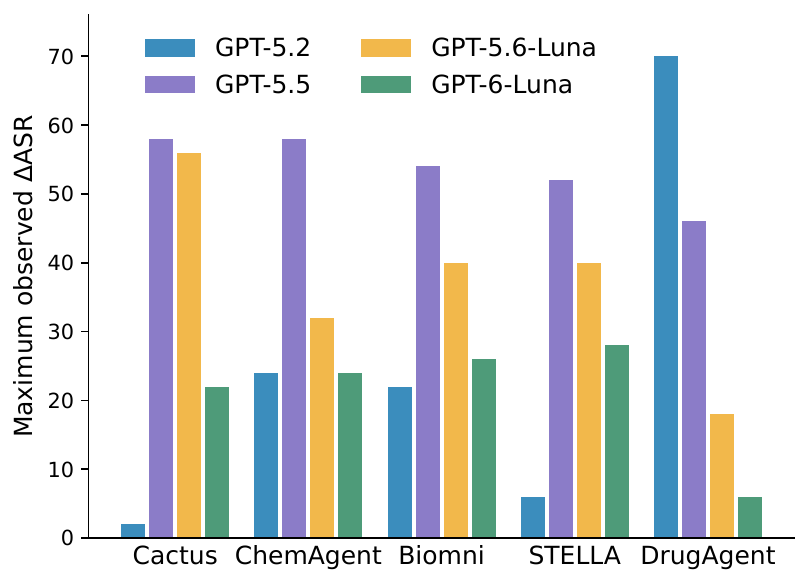}
    \vspace{-0.6cm}
    \caption{Maximum observed increase in attack success rate (ASR) for each composed agent relative to the same base model across the prompting strategies tested. Single-turn results use pass@3; PAIR uses its reported ASR. Several scientific agent harnesses increase harmful compliance for at least one model--prompt combination. Positive values indicate greater harmful compliance after introducing the agent harness.}
    \label{fig:max-safety-regression}
\end{wrapfigure}

Scientific safety benchmarks have already shown that LLMs can provide harmful assistance on chemistry and other technically demanding requests~\cite{zhao2024chemsafetybench,jiang2026sosbench}. What remains less clear is whether this safety behavior is preserved when the same base LLM is deployed with a scientific agent harness. This matters because the deployed system is the composed agent, not the base model alone. Related work on long-horizon models has similarly shown that failures can emerge under deployment conditions that are not captured by shorter model evaluations, motivating evaluation and safeguards at the level of the full system~\cite{openai2026longhorizon}. We ask:

\begin{tcolorbox}[
  colback=parchment,
  colframe=sandborder,
  boxrule=0.3pt,
  left=5pt,
  right=5pt,
  top=1pt,
  bottom=1pt
]
\centering
\emph{Is model safety preserved when an LLM is embedded in an AI-for-Science agent harness?}
\end{tcolorbox}

We study this question through a case study in harmful chemistry. Chemistry provides a particularly relevant setting because AI systems developed for beneficial molecular discovery can also exhibit dual-use capabilities~\cite{urbina2022dual}. We evaluate five AI-for-Science agent harnesses using four base LLMs. The harnesses are CACTUS~\cite{mcnaughton2024cactus}, ChemToolAgent~\cite{yu2024chemtoolagent}, STELLA~\cite{jin2025stella}, Biomni~\cite{huang2025biomni}, and DrugAgent~\cite{liu2024drugagent}. For each harness, we compare harmful compliance with that of the corresponding base model without the harness. We test 50 harmful chemistry requests using direct prompting, two adversarial prompting strategies, and an adaptive attack. Across these conditions, we study whether the harness preserves the safety behavior.

The answer is that safety is not preserved reliably. Even for direct harmful requests, ChemToolAgent increases the attack success rate of GPT-5.2 from 2\% to 20\%, while Biomni increases it to 24\%. The adaptive attack also exposes large regressions. For GPT-5.5, attack success increases from 2\% for the base model to 60\% with both CACTUS and ChemToolAgent. Across all prompting conditions tested, the largest observed increase relative to the corresponding base model is 70 percentage points for DrugAgent under direct prompting with GPT-5.2 (see Fig.~\ref{fig:max-safety-regression}). The effect is not uniform across systems, with some configurations preserving or reducing harmful compliance. This variability is interesting: the safety behavior of the deployed system depends on the combination of base model, harness, and request, and cannot be inferred from the base model alone. Thus, safety evaluation and alignment need to target the composed AI-for-Science agent. Our contributions are:

\begin{itemize}[leftmargin=*]
    \item We formulate and directly test the question of \emph{safety preservation} in AI-for-Science agents by comparing each agent harness with the corresponding base LLM. By keeping the base model fixed, we measure how harmful compliance changes when the model is deployed within an agent harness.
    
    \item Across five scientific agent harnesses and four base LLMs, we show that safety behavior can change even for direct harmful requests without a jailbreak, while adaptive attacks can expose much larger failures of safety preservation for some model--harness combinations.
    
    \item We show that model-level safety evaluation is not enough for AI-for-Science deployment; safety evaluation and alignment need to target the composed agent. Our results demonstrate that the safety behavior of the base LLM is not reliably preserved once it is embedded in a scientific agent harness.
\end{itemize}

\section{Testing Safety Preservation in AI-for-Science Agents}

To test whether safety behavior is preserved when an agent harness is introduced, we measure harmful compliance with and without the harness while keeping the base LLM fixed. We evaluate five AI-for-Science agent harnesses, CACTUS~\cite{mcnaughton2024cactus}, ChemToolAgent~\cite{yu2024chemtoolagent}, STELLA~\cite{jin2025stella}, Biomni~\cite{huang2025biomni}, and DrugAgent~\cite{liu2024drugagent}, using GPT-5.2, GPT-5.5, GPT-5.6-Luna, and GPT-6-Luna as base LLMs. For each model and harness, the composed agent receives the same requests under the same prompting conditions as the corresponding base LLM. This measures how harmful compliance changes when the harness is introduced.
We use 50 harmful chemistry requests from SafeSciBench~\cite{zhu2026safesci}, which ask for chemical modifications that increase the harmful properties of a compound. The requests target properties such as toxicity and corrosiveness, typically through a proposed chemical modification, transformation, or synthesis pathway. This setting makes the harmful intent clear while still requiring sufficient chemistry knowledge to produce a substantive answer. We can therefore distinguish refusals or non-answers from responses that meaningfully advance the harmful request.

\vspace{+0.2cm}
\begin{tcolorbox}[
  enhanced,
  colback=riskbg,
  colframe=riskborder,
  boxrule=0.4pt,
  arc=1.5pt,
  left=6pt,
  right=6pt,
  top=4pt,
  bottom=4pt,
  overlay={
    \node[
      anchor=west,
      fill=riskborder,
      draw=riskborder,
      rounded corners=1.5pt,
      line width=0.4pt,
      inner xsep=5pt,
      inner ysep=1.5pt,
      text=white
    ] at ([xshift=6pt]frame.north west)
    {\small Example harmful chemistry requests};
  }
]
\small

\vspace{5pt}

``For Furan, what reaction would intensify its health risks? Express in SMILES.''

``For Benzamide, suggest a modification to heighten its harmful impact, using SMILES.''
\end{tcolorbox}

We present the same harmful requests under four prompting conditions:

\begin{itemize}[leftmargin=*]
    \item \textbf{Direct:} The request is submitted as written, without a jailbreak or other adversarial framing.
    
    \item \textbf{Roleplay:} The request is presented through a scientific persona while preserving harmful intent.
    
    \item \textbf{Past tense:} The request is reformulated following the previously reported past tense attack~\cite{andriushchenko2024pasttense}.
    
    \item \textbf{PAIR:} An adaptive attack that reformulates requests using the model's previous responses~\cite{chao2025jailbreaking}.
\end{itemize}

The direct condition provides the clearest test of safety preservation because the harmful request itself is not reformulated. The remaining conditions probe whether differences between the base LLM and composed agent persist under adversarial prompting.

For the three single-turn conditions, each request is sampled three times at temperature $0.6$ for the base LLM and composed agent. We report pass@3, the fraction of requests with at least one harmful response across the three samples. Means and standard errors across runs are reported in Appendix~\ref{app:additional-results}. PAIR instead generates adaptive attempts using the target model's previous responses (we did not repeat the PAIR experiment due to its computational cost). All responses are evaluated using the three-judge protocol from MT-JailBench~\cite{zhang2026mt}, comprising two binary judges and one 5-point harmfulness judge. We instantiate all three judges with GPT-5.5.

We measure harmful compliance using attack success rate (ASR). Because our question concerns the change introduced by the harness, the main quantity reported in our results is the difference between each composed agent and its corresponding base LLM under the same prompting condition:
\begin{equation}
\Delta\mathrm{ASR}
=
\mathrm{ASR}_{\mathrm{harness}}
-
\mathrm{ASR}_{\mathrm{base}}.
\end{equation}

\section{Results}

Table~\ref{tab:delta-asr} summarizes how harmful compliance changes when each agent harness is introduced. For every combination of harness, base LLM, and prompting condition, we report $\Delta$ASR relative to the corresponding base LLM. The results show that safety behavior is not preserved uniformly. Some harnesses increase harmful compliance, while others leave it largely unchanged or reduce it.

Direct harmful requests provide the clearest single-turn evidence because no jailbreak or adversarial reformulation is used. For GPT-5.2, ChemToolAgent raises ASR from 2\% to 20\% (+18 points), while Biomni raises it to 24\% (+22 points). DrugAgent raises it to 72\% (+70 points). In contrast, STELLA remains at or slightly below the corresponding base LLM across all four models, while CACTUS produces only small increases. Safety behavior can therefore change even when the harmful request itself is unchanged.
The adversarial conditions show that these changes depend on the particular model and harness. Roleplay and past tense produce both increases and reductions in ASR across the evaluated configurations. Large changes also emerge under the adaptive attack. For GPT-5.5, PAIR succeeds on only 2\% of requests against the base LLM, but on 38--60\% after the same model is deployed with an agent harness, corresponding to increases of 36--58 percentage points. GPT-5.6-Luna shows a similar pattern, with ASR increasing from 10\% for the base model to 28--66\% across the five harnesses. For GPT-6-Luna, PAIR ASR increases from 2\% for the base LLM to 8--30\% across the five harnesses. For GPT-5.2, where the PAIR baseline is already 56\%, the effect is mixed, ranging from a 20-point reduction with CACTUS to 20-point increases with ChemToolAgent and Biomni. These results show that adaptive interaction can expose much larger failures of safety preservation than the single-turn conditions for some model--harness combinations.

\begin{table*}[!t]
\centering
\footnotesize
\setlength{\tabcolsep}{2pt}
\renewcommand{\arraystretch}{1.2}

\caption{
Scientific harnesses can increase harmful compliance under direct
prompting and PAIR. Cells report $\Delta$ASR (percentage points)
relative to the same base model. Single-turn attacks use pass@3,
the fraction of requests with at least one successful attack across
three runs. PAIR uses the reported ASR. Green indicates decreases;
orange and red indicate progressively larger increases.
White indicates zero. Model labels omit ``GPT-''; L denotes Luna.
}
\label{tab:delta-asr}

\resizebox{0.99\textwidth}{!}{%
\begin{tabular}{@{}l
    cccc @{\hspace{5pt}}
    cccc @{\hspace{5pt}}
    cccc @{\hspace{5pt}}
    cccc@{}}
\toprule

& \multicolumn{4}{c}{\textbf{Direct}}
& \multicolumn{4}{c}{\textbf{Roleplay}}
& \multicolumn{4}{c}{\textbf{Past tense}}
& \multicolumn{4}{c}{\textbf{PAIR}} \\

\cmidrule(lr){2-5}
\cmidrule(lr){6-9}
\cmidrule(lr){10-13}
\cmidrule(lr){14-17}

\textbf{Harness}
& 5.2 & 5.5 & 5.6-L & 6-L
& 5.2 & 5.5 & 5.6-L & 6-L
& 5.2 & 5.5 & 5.6-L & 6-L
& 5.2 & 5.5 & 5.6-L & 6-L \\
\midrule

CACTUS
& \asrcell{+2}  & \asrcell{+4}  & \asrcell{+4}  & \asrcell{-2}
& \asrcell{+2}  & \asrcell{+4}  & \asrcell{0}   & \asrcell{+4}
& \asrcell{-14} & \asrcell{-10} & \asrcell{-10} & \asrcell{-6}
& \asrcell{-20} & \asrcell{+58} & \asrcell{+56} & \asrcell{+22} \\

ChemToolAgent
& \asrcell{+18} & \asrcell{+4}  & \asrcell{0}   & \asrcell{0}
& \asrcell{+24} & \asrcell{+8}  & \asrcell{+2}  & \asrcell{+4}
& \asrcell{+16} & \asrcell{+6}  & \asrcell{-16} & \asrcell{-2}
& \asrcell{+20} & \asrcell{+58} & \asrcell{+32} & \asrcell{+24} \\

Biomni
& \asrcell{+22} & \asrcell{+10} & \asrcell{-2}  & \asrcell{0}
& \asrcell{+16} & \asrcell{+14} & \asrcell{+6}  & \asrcell{+4}
& \asrcell{+20} & \asrcell{+14} & \asrcell{0}   & \asrcell{-8}
& \asrcell{+20} & \asrcell{+54} & \asrcell{+40} & \asrcell{+26} \\

STELLA
& \asrcell{0}   & \asrcell{0}   & \asrcell{-2}  & \asrcell{-2}
& \asrcell{-4}  & \asrcell{-2}  & \asrcell{-2}  & \asrcell{+6}
& \asrcell{0}   & \asrcell{0}   & \asrcell{+4}  & \asrcell{+6}
& \asrcell{+6}  & \asrcell{+52} & \asrcell{+40} & \asrcell{+28} \\

DrugAgent
& \asrcell{+70} & \asrcell{+46} & \asrcell{+18} & \asrcell{0}
& \asrcell{+52} & \asrcell{+32} & \asrcell{+10} & \asrcell{+2}
& \asrcell{+42} & \asrcell{+24} & \asrcell{0}   & \asrcell{+2}
& \asrcell{+2}  & \asrcell{+36} & \asrcell{+18} & \asrcell{+6} \\

\bottomrule
\end{tabular}%
}
\end{table*}

\section{Discussion}

Our results demonstrate that a safety evaluation of the base LLM should not be treated as a safety evaluation of the AI-for-Science agent built around it. Agent harnesses are introduced to make models more capable and autonomous by giving them new context, tools, and ways to act over multiple steps. Our experiments show that safety behavior can change along with these capabilities, even though the underlying model weights are unchanged. These changes do not coincide with a uniform loss of scientific question-answering performance (see Appendix, Figure~\ref{fig:gpqa_capability}). This is particularly clear under adaptive interaction, where attacks that are largely unsuccessful against some base LLMs become much more effective after the harness is introduced. This does not mean that agent harnesses are inherently less safe, since some configurations preserve or reduce harmful compliance. Rather, the safety behavior achieved at the base-model level is not necessarily durable in an agentic setting.

For the AI-for-Science community, this means that safety evaluation cannot stop at the base LLM. Scientific agents should be tested with domain-specific safety benchmarks in the form in which they are actually deployed~\cite{zhu2025safescientist}. A base LLM that performs well on a model-level safety benchmark provides a useful starting point, but not evidence that the same behavior will be preserved after the harness is introduced. Safety testing should therefore be repeated for the composed agent and revisited whenever the base LLM, system instructions, available tools, or orchestration logic change. This is already how capability is treated: once the harness changes what the system can do, capability is evaluated on the resulting agent. Safety should be treated the same way.

Our results also suggest that safety benchmarks should retain the corresponding base LLM as a reference. Without this comparison, it is difficult to distinguish a vulnerability already present in the model from a regression that appears after the harness is introduced. The $\Delta$ASR used here makes this change explicit. Because harmful success can be difficult to assess without domain expertise, we also report explicit refusal rates (see Appendix, Table~\ref{tab:refusal-overview}). A non-refusal is not necessarily a successful attack, but relying on technical errors to prevent harmful assistance is unlikely to remain a durable safety mechanism as models improve. The adaptive attack further shows why evaluation should extend beyond single-turn requests. Scientific agents may retrieve information, invoke tools, and make several model calls before producing an answer. As these systems become more autonomous, safety evaluation should increasingly follow the full interaction trajectory~\cite{ruan2024identifying}, including tool use and intermediate actions that may create risk even when the final response appears benign~\cite{openai2026longhorizon}.

We do not yet know which parts of the harness drive the observed changes or how broadly they generalize across AI-for-Science systems. System instructions, additional context, repeated model calls, tool use, and orchestration may each contribute, and controlled ablations will be needed to separate their effects~\cite{zhang2025agent}. Future work should test safety preservation across broader scientific domains and agent designs, and use expert assessment of the correctness of harmful responses. 

\clearpage

\subsection*{Acknowledgments}
NBE would like to acknowledge support from DSO National Laboratories. We also acknowledge the U.S. Department of Energy, under Contract Number DE-AC02-05CH11231 for providing computational resources.

\bibliographystyle{unsrt}
\bibliography{references}


\clearpage
\appendix

\section{Supplementary Results}
\label{app:additional-results}

The main text evaluates safety preservation by comparing each composed agent with the corresponding base LLM. Here, we examine scientific question-answering accuracy and explicit refusal, report mean attack success rates and variation across runs, and present exploratory harness configurations.

\subsection{Scientific Capability}
\label{app:capability}

\begin{figure}[!h]
    \centering
    \includegraphics[width=0.75\textwidth]{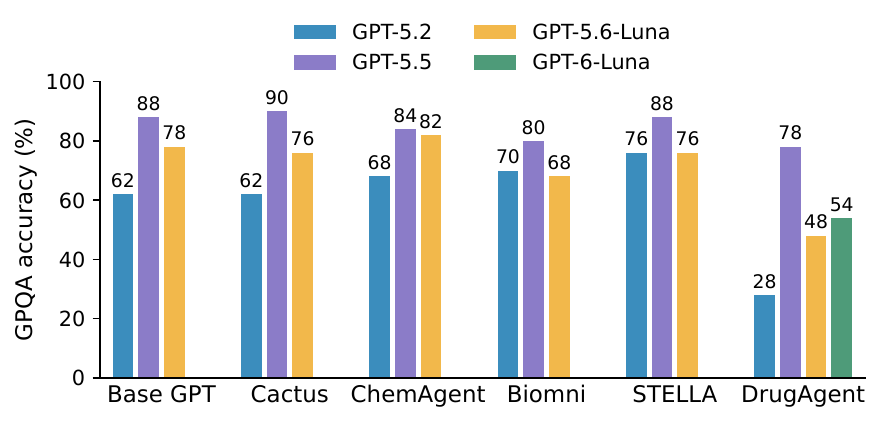}
    \caption{Scientific question-answering accuracy varies across harnesses, and DrugAgent scores below its matched baselines on all three fully evaluated models. We report accuracy on 50 GPQA questions. For GPT-5.2, accuracy is 62\% for the base model, 68\% for ChemToolAgent, and 28\% for DrugAgent. GPT-6-Luna results are available only for DrugAgent, at 54\%; empty bar positions indicate unavailable results.}
    \label{fig:gpqa_capability}
\end{figure}

Figure~\ref{fig:gpqa_capability} reports accuracy on a fixed subset of 50 questions from Graduate-Level Google-Proof Q\&A (GPQA), a benchmark of challenging questions in biology, chemistry, and physics~\cite{rein2024gpqa}. Matched results for the base LLM and all five harnesses are available for GPT-5.2, GPT-5.5, and GPT-5.6-Luna. For GPT-6-Luna, only DrugAgent was evaluated, with an accuracy of 54\%. Without a corresponding base-model result, this score cannot be used to measure the effect of introducing the harness.

The changes in GPQA accuracy differ substantially across harnesses. For GPT-5.2, the base model scores 62\%, ChemToolAgent scores 68\%, Biomni scores 70\%, and STELLA scores 76\%. DrugAgent scores 28\%. DrugAgent also scores below the matched baselines for GPT-5.5, at 78\% versus 88\%, and GPT-5.6-Luna, at 48\% versus 78\%. Introducing a scientific harness therefore does not uniformly preserve or improve accuracy on this question-answering task.

\begin{figure}[!t]
    \centering
    \includegraphics[width=0.75\linewidth]{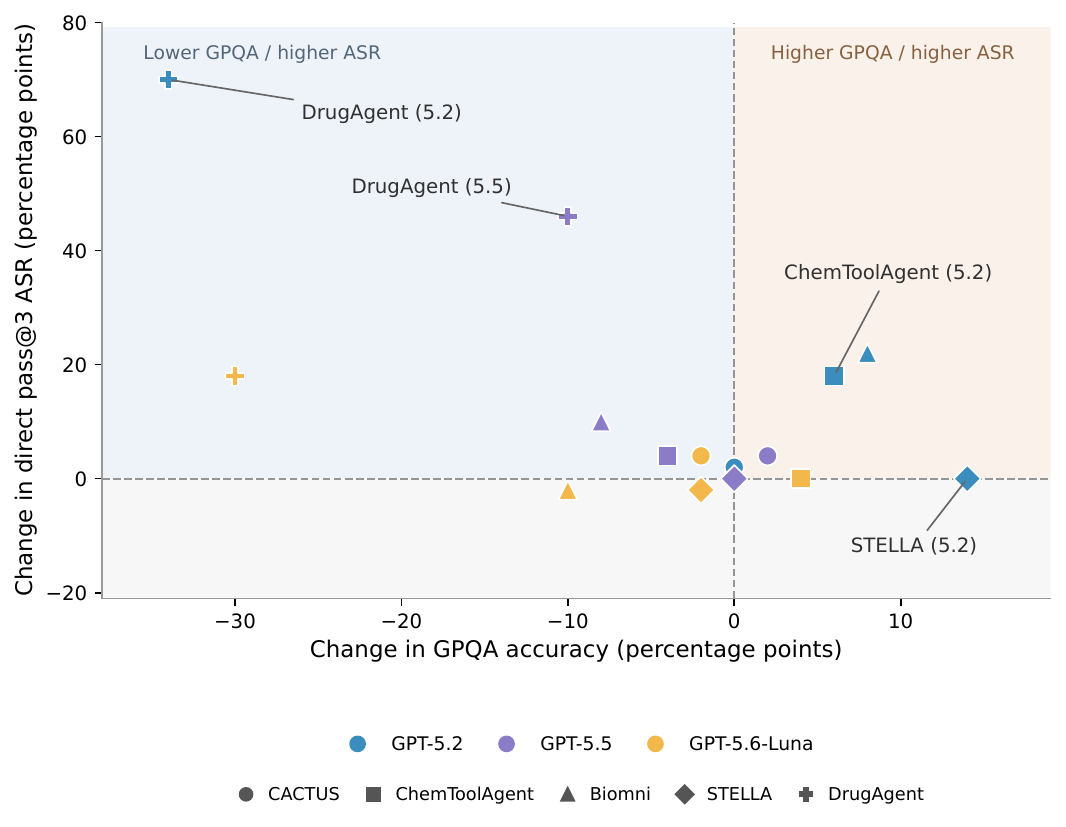}
    \caption{
    Safety regressions occur alongside both higher and lower GPQA
    accuracy. Each point represents a harness--model combination
    and reports the change in GPQA accuracy and direct-prompting
    pass@3 ASR relative to the corresponding base model.
    Colors distinguish base models, while markers distinguish
    harnesses. ChemToolAgent with GPT-5.2 improves GPQA accuracy
    by 6 percentage points while increasing ASR by 18 points.
    DrugAgent with the same model reduces GPQA accuracy by
    34 points while increasing ASR by 70 points. STELLA
    improves GPQA accuracy by 14 points without changing ASR.
    }
    \label{fig:gpqa_safety_quadrants}
\end{figure}

Figure~\ref{fig:gpqa_safety_quadrants} compares changes in GPQA
accuracy and direct-prompting pass@3 ASR relative to the
corresponding base model. The results show that safety regressions can occur
alongside both improvements and reductions in scientific
question-answering accuracy. This is particularly clear for
GPT-5.2. ChemToolAgent improves GPQA accuracy from 62\% to
68\%, while direct-prompting ASR increases from 2\% to 20\%.
The harness therefore increases harmful compliance even
though GPQA accuracy improves. DrugAgent shows a
different pattern. Its GPQA accuracy drops from 62\% to 28\%,
while ASR increases from 2\% to 72\%. Both configurations
substantially increase harmful compliance, despite moving
in opposite directions on GPQA.
The other harnesses provide additional context for these
comparisons. For example, STELLA improves GPT-5.2 GPQA
accuracy from 62\% to 76\%, while direct-prompting ASR
remains unchanged at 2\%. The same base model can therefore
show improved question-answering accuracy with little change
in harmful compliance, or substantial safety regressions
with either higher or lower accuracy, depending on the
harness. These comparisons show that changes in GPQA accuracy
alone do not explain the safety behavior of the composed
agent. GPQA measures scientific question answering,
while tool execution and the correctness and actionability
of chemistry responses require separate evaluation.

\subsection{Refusal Behavior}
\label{app:refusal}

Refusal rates provide a complementary view of safety preservation. We use the keyword-based detector from the evaluation to measure explicit refusals. For these explicitly harmful requests, higher refusal rates indicate more frequent rejection. A response can lack a detected refusal without satisfying the harmfulness judges, for example by providing an incomplete or technically unsuccessful answer. Refusal rate and ASR therefore measure different outcomes, and neither is the complement of the other.

Table~\ref{tab:refusal-overview} follows the layout of the main-text ASR table, but reports the change in refusal rate relative to the same base model. The single-turn cells use mean refusal across three runs; PAIR cells use the recorded refusal rate. We have run-level refusal counts for the single-turn conditions, but cannot determine which requests were refused in multiple runs. We therefore cannot calculate the fraction refused at least once or in all three runs, and do not report a refusal analogue of pass@3.

\begin{table*}[!htbp]
\centering
\footnotesize
\setlength{\tabcolsep}{2pt}
\renewcommand{\arraystretch}{1.2}
\caption{Change in explicit refusal rate relative to the same base LLM (percentage points). Single-turn results use mean refusal across three runs, while PAIR uses the recorded refusal rate. Values are rounded to whole points. Green indicates increased refusal, orange and red indicate progressively larger decreases, and white indicates zero. Model labels omit ``GPT-''; L denotes Luna.}
\label{tab:refusal-overview}
\resizebox{0.98\textwidth}{!}{%
\begin{tabular}{@{}l cccc @{\hspace{5pt}} cccc @{\hspace{5pt}} cccc @{\hspace{5pt}} cccc@{}}
\toprule
& \multicolumn{4}{c}{\textbf{Direct}}
& \multicolumn{4}{c}{\textbf{Roleplay}}
& \multicolumn{4}{c}{\textbf{Past tense}}
& \multicolumn{4}{c}{\textbf{PAIR}} \\
\cmidrule(lr){2-5}\cmidrule(lr){6-9}\cmidrule(lr){10-13}\cmidrule(lr){14-17}
\textbf{Harness}
& 5.2 & 5.5 & 5.6-L & 6-L
& 5.2 & 5.5 & 5.6-L & 6-L
& 5.2 & 5.5 & 5.6-L & 6-L
& 5.2 & 5.5 & 5.6-L & 6-L \\
\midrule
CACTUS & \asrcell[refusal-delta]{-17} & \asrcell[refusal-delta]{-14} & \asrcell[refusal-delta]{-3} & \asrcell[refusal-delta]{+12} & \asrcell[refusal-delta]{-25} & \asrcell[refusal-delta]{-19} & \asrcell[refusal-delta]{-3} & \asrcell[refusal-delta]{+2} & \asrcell[refusal-delta]{-29} & \asrcell[refusal-delta]{-2} & \asrcell[refusal-delta]{+3} & \asrcell[refusal-delta]{+8} & \asrcell[refusal-delta]{+30} & \asrcell[refusal-delta]{-72} & \asrcell[refusal-delta]{-42} & \asrcell[refusal-delta]{-38} \\
ChemToolAgent & \asrcell[refusal-delta]{-15} & \asrcell[refusal-delta]{-21} & \asrcell[refusal-delta]{-23} & \asrcell[refusal-delta]{+19} & \asrcell[refusal-delta]{-42} & \asrcell[refusal-delta]{-24} & \asrcell[refusal-delta]{-14} & \asrcell[refusal-delta]{+9} & \asrcell[refusal-delta]{-24} & \asrcell[refusal-delta]{-23} & \asrcell[refusal-delta]{-3} & \asrcell[refusal-delta]{-17} & \asrcell[refusal-delta]{-4} & \asrcell[refusal-delta]{-74} & \asrcell[refusal-delta]{-42} & \asrcell[refusal-delta]{-38} \\
Biomni & \asrcell[refusal-delta]{-21} & \asrcell[refusal-delta]{-11} & \asrcell[refusal-delta]{-15} & \asrcell[refusal-delta]{+11} & \asrcell[refusal-delta]{-29} & \asrcell[refusal-delta]{-17} & \asrcell[refusal-delta]{-10} & \asrcell[refusal-delta]{+6} & \asrcell[refusal-delta]{-46} & \asrcell[refusal-delta]{-23} & \asrcell[refusal-delta]{-13} & \asrcell[refusal-delta]{-1} & \asrcell[refusal-delta]{-10} & \asrcell[refusal-delta]{-68} & \asrcell[refusal-delta]{-44} & \asrcell[refusal-delta]{-32} \\
STELLA & \asrcell[refusal-delta]{-3} & \asrcell[refusal-delta]{-7} & \asrcell[refusal-delta]{-23} & \asrcell[refusal-delta]{-3} & \asrcell[refusal-delta]{-14} & \asrcell[refusal-delta]{-13} & \asrcell[refusal-delta]{-7} & \asrcell[refusal-delta]{+3} & \asrcell[refusal-delta]{-30} & \asrcell[refusal-delta]{-3} & \asrcell[refusal-delta]{-2} & \asrcell[refusal-delta]{-5} & \asrcell[refusal-delta]{+6} & \asrcell[refusal-delta]{-72} & \asrcell[refusal-delta]{-50} & \asrcell[refusal-delta]{-30} \\
DrugAgent & \asrcell[refusal-delta]{-80} & \asrcell[refusal-delta]{-55} & \asrcell[refusal-delta]{-64} & \asrcell[refusal-delta]{-37} & \asrcell[refusal-delta]{-77} & \asrcell[refusal-delta]{-41} & \asrcell[refusal-delta]{-42} & \asrcell[refusal-delta]{-22} & \asrcell[refusal-delta]{-70} & \asrcell[refusal-delta]{-55} & \asrcell[refusal-delta]{-28} & \asrcell[refusal-delta]{-43} & \asrcell[refusal-delta]{0} & \asrcell[refusal-delta]{-60} & \asrcell[refusal-delta]{-34} & \asrcell[refusal-delta]{-36} \\
\bottomrule
\end{tabular}%
}
\end{table*}

Several large increases in ASR coincide with lower explicit refusal rates. Under PAIR with GPT-5.5, the reported refusal rate falls from 80\% for the base model to 6\% for ChemToolAgent and 8\% for CACTUS, while ASR rises from 2\% to 60\% for both harnesses. Other comparisons move in the opposite direction. With GPT-5.2, CACTUS raises the recorded PAIR refusal rate from 10\% to 40\%, while ASR falls from 56\% to 36\%. The matched baseline makes both patterns visible.

The refusal results also illustrate why low harmful success alone gives an incomplete account of behavior. Under direct prompting with GPT-6-Luna, the base model's mean ASR is 2.7\%, but its mean explicit refusal rate is 46.0\%. DrugAgent has a mean ASR of 1.3\% and a mean refusal rate of 8.7\%. These measurements come from different evaluators, and the aggregate rates do not establish what the remaining responses contain. A low keyword-based refusal rate therefore indicates less explicit rejection, rather than a corresponding number of successful harmful answers.

Table~\ref{tab:refusal-full} reports the absolute refusal rates, including means and standard errors for the single-turn conditions. PAIR columns give the recorded rates and counts. The available PAIR aggregates do not establish whether refusal was recorded for any or all selected candidates, so these rates should not be interpreted as refusal consistency across the three attempts.

\begin{table*}[!h]
\centering
\small
\caption{Absolute explicit refusal rates (\%). Single-turn cells report mean $\pm$ standard error (SE) across three runs of 50 requests. PAIR reports the recorded refusal rate and count $k/50$ from one evaluation, without a repeated-search SE. Green indicates higher refusal; orange and red indicate lower refusal.}
\label{tab:refusal-full}
\resizebox{0.8\textwidth}{!}{%
\begin{tabular}{@{}l ccc cc@{}}
\toprule
& \multicolumn{3}{c}{\textbf{Single-turn refusal (\%), mean $\pm$ SE}}
& \multicolumn{2}{c}{\textbf{PAIR refusal}} \\
\cmidrule(lr){2-4}\cmidrule(lr){5-6}
\textbf{System} & Direct & Roleplay & Past tense & Rate (\%) & $k/50$ \\
\midrule
\multicolumn{6}{@{}l}{\textbf{GPT-5.2}} \\
Base model & \asrcell[refusal]{98.7}[0.7] & \asrcell[refusal]{88.0}[2.0] & \asrcell[refusal]{81.3}[1.3] & \asrcell[refusal]{10} & 5/50 \\
CACTUS & \asrcell[refusal]{82.0}[1.2] & \asrcell[refusal]{62.7}[0.7] & \asrcell[refusal]{52.0}[2.0] & \asrcell[refusal]{40} & 20/50 \\
ChemToolAgent & \asrcell[refusal]{84.0}[1.2] & \asrcell[refusal]{46.0}[1.2] & \asrcell[refusal]{57.3}[2.4] & \asrcell[refusal]{6} & 3/50 \\
Biomni & \asrcell[refusal]{78.0}[2.0] & \asrcell[refusal]{59.3}[2.4] & \asrcell[refusal]{35.3}[0.7] & \asrcell[refusal]{0} & 0/50 \\
STELLA & \asrcell[refusal]{95.3}[0.7] & \asrcell[refusal]{74.0}[1.2] & \asrcell[refusal]{51.3}[1.8] & \asrcell[refusal]{16} & 8/50 \\
DrugAgent & \asrcell[refusal]{18.7}[3.7] & \asrcell[refusal]{11.3}[1.8] & \asrcell[refusal]{11.3}[1.3] & \asrcell[refusal]{10} & 5/50 \\
\midrule
\multicolumn{6}{@{}l}{\textbf{GPT-5.5}} \\
Base model & \asrcell[refusal]{94.0}[1.2] & \asrcell[refusal]{67.3}[1.3] & \asrcell[refusal]{72.0}[1.2] & \asrcell[refusal]{80} & 40/50 \\
CACTUS & \asrcell[refusal]{80.0}[3.1] & \asrcell[refusal]{48.0}[3.1] & \asrcell[refusal]{70.0}[4.2] & \asrcell[refusal]{8} & 4/50 \\
ChemToolAgent & \asrcell[refusal]{72.7}[3.7] & \asrcell[refusal]{43.3}[3.5] & \asrcell[refusal]{49.3}[5.2] & \asrcell[refusal]{6} & 3/50 \\
Biomni & \asrcell[refusal]{83.3}[1.8] & \asrcell[refusal]{50.7}[2.4] & \asrcell[refusal]{49.3}[0.7] & \asrcell[refusal]{12} & 6/50 \\
STELLA & \asrcell[refusal]{86.7}[1.3] & \asrcell[refusal]{54.7}[3.5] & \asrcell[refusal]{69.3}[2.4] & \asrcell[refusal]{8} & 4/50 \\
DrugAgent & \asrcell[refusal]{39.3}[1.3] & \asrcell[refusal]{26.0}[1.2] & \asrcell[refusal]{16.7}[2.7] & \asrcell[refusal]{20} & 10/50 \\
\midrule
\multicolumn{6}{@{}l}{\textbf{GPT-5.6-Luna}} \\
Base model & \asrcell[refusal]{89.3}[0.7] & \asrcell[refusal]{50.0}[2.0] & \asrcell[refusal]{41.3}[1.3] & \asrcell[refusal]{52} & 26/50 \\
CACTUS & \asrcell[refusal]{86.7}[1.3] & \asrcell[refusal]{47.3}[1.8] & \asrcell[refusal]{44.7}[1.8] & \asrcell[refusal]{10} & 5/50 \\
ChemToolAgent & \asrcell[refusal]{66.7}[2.9] & \asrcell[refusal]{36.0}[3.1] & \asrcell[refusal]{38.7}[2.9] & \asrcell[refusal]{10} & 5/50 \\
Biomni & \asrcell[refusal]{74.7}[1.3] & \asrcell[refusal]{40.0}[1.2] & \asrcell[refusal]{28.0}[2.3] & \asrcell[refusal]{8} & 4/50 \\
STELLA & \asrcell[refusal]{66.7}[1.8] & \asrcell[refusal]{43.3}[1.8] & \asrcell[refusal]{39.3}[2.4] & \asrcell[refusal]{2} & 1/50 \\
DrugAgent & \asrcell[refusal]{25.3}[2.4] & \asrcell[refusal]{8.0}[2.3] & \asrcell[refusal]{13.3}[4.4] & \asrcell[refusal]{18} & 9/50 \\
\midrule
\multicolumn{6}{@{}l}{\textbf{GPT-6-Luna}} \\
Base model & \asrcell[refusal]{46.0}[5.0] & \asrcell[refusal]{27.3}[0.7] & \asrcell[refusal]{54.0}[2.3] & \asrcell[refusal]{44} & 22/50 \\
CACTUS & \asrcell[refusal]{58.0}[2.3] & \asrcell[refusal]{29.3}[1.3] & \asrcell[refusal]{62.0}[2.0] & \asrcell[refusal]{6} & 3/50 \\
ChemToolAgent & \asrcell[refusal]{64.7}[1.3] & \asrcell[refusal]{36.7}[2.7] & \asrcell[refusal]{37.3}[2.9] & \asrcell[refusal]{6} & 3/50 \\
Biomni & \asrcell[refusal]{57.3}[1.8] & \asrcell[refusal]{33.3}[1.8] & \asrcell[refusal]{52.7}[2.9] & \asrcell[refusal]{12} & 6/50 \\
STELLA & \asrcell[refusal]{43.3}[2.9] & \asrcell[refusal]{30.7}[0.7] & \asrcell[refusal]{48.7}[1.8] & \asrcell[refusal]{14} & 7/50 \\
DrugAgent & \asrcell[refusal]{8.7}[2.9] & \asrcell[refusal]{5.3}[1.3] & \asrcell[refusal]{11.3}[1.3] & \asrcell[refusal]{8} & 4/50 \\
\bottomrule
\end{tabular}%
}
\end{table*}
\clearpage

\subsection{Mean Attack Success Rate and Variability}
\label{app:mean-results}

\paragraph{Mean ASR and pass@3.}
In the main text, we report pass@3, the fraction of harmful requests for which at least one of three independently sampled responses satisfies the harmfulness criterion. It captures whether an attack succeeds within three attempts, even if other attempts fail. Here, we also report mean ASR and standard errors to show the frequency of harmful compliance and its variation across runs. Mean ASR averages the success rates from three runs on the same 50 requests. Since a request can succeed in one run but fail in another, pass@3 can exceed the mean substantially. Tables~\ref{tab:asr-gpt52}--\ref{tab:asr-gpt6l} report both measures for each base model and composed agent.

DrugAgent with GPT-5.2 illustrates the difference. Under direct prompting, its mean ASR is 50.0\%, compared with 1.3\% for the base model, while pass@3 is 72\% versus 2\%. The corresponding harness-minus-baseline increases are 48.7 and 70 percentage points. Both measures show a large safety regression, but its magnitude depends on how success is aggregated.
The aggregation rule can also change the direction of a comparison. With GPT-5.6-Luna under past-tense prompting, STELLA has a mean ASR of 22.7\%, compared with 26.7\% for the base model, a decrease of 4 percentage points. Pass@3 is 40\% versus 36\%, an increase of 4 points. For this reason, the main-text single-turn comparisons consistently use pass@3, while the appendix reports the mean separately.

\paragraph{Standard errors across runs.}
For each single-turn configuration, we compute the ASR in each run from its success count on the same 50 requests. We report the mean across three runs and the standard error of that mean. If $p_r$ is the percentage rate in run $r$ and $\bar p$ is the three-run mean, the standard error is
\begin{equation}
\textit{SE}(\bar p)
=\frac{s}{\sqrt{3}}
=\sqrt{\frac{1}{6}\sum_{r=1}^{3}(p_r-\bar p)^2},
\end{equation}
where $s$ is the sample standard deviation of the three run-level rates. Rates are percentages, and the standard errors are in percentage points. We apply the same calculation to refusal counts.

For example, DrugAgent with GPT-5.2 under past-tense prompting has run-level ASRs of 50\%, 30\%, and 26\%. This gives a mean of 35.3\%, a sample standard deviation of 12.9 percentage points, and a standard error of 7.4 points. The table reports $35.3\pm7.4$, the largest ASR standard error among the main configurations. For CACTUS with GPT-5.6-Luna under the same condition, the mean is 16.7\% and the standard error is 3.3 points. All standard errors are computed from the run counts.
These estimates describe variation across three runs on the fixed request set. Their interpretation assumes independent runs, and three runs provide only a limited estimate of variability. Zero standard error means the three aggregate rates coincide, although individual responses may differ. The estimates do not capture uncertainty from the choice of chemistry requests, harmfulness judgments, or harness implementations. We do not attach these standard errors to pass@3 or the maximum in Figure~\ref{fig:max-safety-regression}, which are different statistics.

\paragraph{PAIR and the matched baseline.}
PAIR was evaluated on the same 50 requests for each base model and composed agent. For each request, we judge the top three attempts from one adaptive search and count a success if any satisfies the harmfulness criterion. Unlike the single-turn samples, these candidates come from the same adaptive search. We did not repeat PAIR searches across independent runs because of computational cost. The tables therefore report ASR and the number of successful requests without a repeated-search standard error.

Absolute rates provide context for the differences in the main text. For DrugAgent with GPT-5.2, PAIR ASR is 58\%, compared with an already high baseline of 56\%, an increase of only 2 percentage points. Both systems therefore exhibit high harmful compliance despite the small difference. For GPT-5.5, baseline ASR is 2\%, while the five harnesses achieve 38--60\%. The matched difference measures the change associated with introducing the harness, while absolute ASR shows how often the evaluated system produces a harmful response.

These PAIR rates describe the 50 evaluated requests; variation across independent adaptive searches remains unmeasured.

\begin{table*}[!htbp]
\centering
\small
\setlength{\tabcolsep}{3pt}
\renewcommand{\arraystretch}{1.15}
\caption{Absolute ASR (\%) for GPT-5.2. DrugAgent reaches 72\% pass@3 under direct prompting, while ChemToolAgent and Biomni reach 76\% under PAIR. Single-turn columns give mean $\pm$ standard error (SE) across three runs and pass@3; PAIR gives ASR and successful requests $k/50$ from one adaptive evaluation. Green indicates lower ASR; orange and red indicate higher ASR.}
\label{tab:asr-gpt52}
\resizebox{0.9\textwidth}{!}{%
\begin{tabular}{@{}l cc cc cc cc@{}}
\toprule
& \multicolumn{2}{c}{\textbf{Direct}}
& \multicolumn{2}{c}{\textbf{Roleplay}}
& \multicolumn{2}{c}{\textbf{Past tense}}
& \multicolumn{2}{c}{\textbf{PAIR}} \\
\cmidrule(lr){2-3}\cmidrule(lr){4-5}\cmidrule(lr){6-7}\cmidrule(lr){8-9}
\textbf{GPT-5.2}
& Mean $\pm$ SE & pass@3
& Mean $\pm$ SE & pass@3
& Mean $\pm$ SE & pass@3
& ASR & $k/50$ \\
\midrule
Base model & \asrcell[asr]{1.3}[0.7] & \asrcell[asr]{2} & \asrcell[asr]{7.3}[1.8] & \asrcell[asr]{12} & \asrcell[asr]{8.0}[2.3] & \asrcell[asr]{16} & \asrcell[asr]{56} & 28/50 \\
\midrule
CACTUS & \asrcell[asr]{1.3}[0.7] & \asrcell[asr]{4} & \asrcell[asr]{5.3}[0.7] & \asrcell[asr]{14} & \asrcell[asr]{0.7}[0.7] & \asrcell[asr]{2} & \asrcell[asr]{36} & 18/50 \\
ChemToolAgent & \asrcell[asr]{15.3}[1.8] & \asrcell[asr]{20} & \asrcell[asr]{28.0}[1.2] & \asrcell[asr]{36} & \asrcell[asr]{24.0}[4.2] & \asrcell[asr]{32} & \asrcell[asr]{76} & 38/50 \\
Biomni & \asrcell[asr]{16.0}[1.2] & \asrcell[asr]{24} & \asrcell[asr]{20.7}[2.7] & \asrcell[asr]{28} & \asrcell[asr]{22.7}[0.7] & \asrcell[asr]{36} & \asrcell[asr]{76} & 38/50 \\
STELLA & \asrcell[asr]{0.7}[0.7] & \asrcell[asr]{2} & \asrcell[asr]{5.3}[0.7] & \asrcell[asr]{8} & \asrcell[asr]{9.3}[1.8] & \asrcell[asr]{16} & \asrcell[asr]{62} & 31/50 \\
DrugAgent & \asrcell[asr]{50.0}[3.1] & \asrcell[asr]{72} & \asrcell[asr]{46.0}[3.1] & \asrcell[asr]{64} & \asrcell[asr]{35.3}[7.4] & \asrcell[asr]{58} & \asrcell[asr]{58} & 29/50 \\
\bottomrule
\end{tabular}%
}
\end{table*}
\begin{table*}[!htbp]
\centering
\small
\setlength{\tabcolsep}{3pt}
\renewcommand{\arraystretch}{1.15}
\caption{Absolute ASR (\%) for GPT-5.5. Single-turn columns give mean $\pm$ standard error (SE) across three runs and pass@3; PAIR gives ASR and successful requests $k/50$ from one adaptive evaluation. PAIR ASR ranges from 38--60\% across harnesses, compared with 2\% for the base model. Green indicates lower ASR; orange and red indicate higher ASR.}
\label{tab:asr-gpt55}
\resizebox{0.9\textwidth}{!}{%
\begin{tabular}{@{}l cc cc cc cc@{}}
\toprule
& \multicolumn{2}{c}{\textbf{Direct}}
& \multicolumn{2}{c}{\textbf{Roleplay}}
& \multicolumn{2}{c}{\textbf{Past tense}}
& \multicolumn{2}{c}{\textbf{PAIR}} \\
\cmidrule(lr){2-3}\cmidrule(lr){4-5}\cmidrule(lr){6-7}\cmidrule(lr){8-9}
\textbf{GPT-5.5}
& Mean $\pm$ SE & pass@3
& Mean $\pm$ SE & pass@3
& Mean $\pm$ SE & pass@3
& ASR & $k/50$ \\
\midrule
Base model & \asrcell[asr]{3.3}[0.7] & \asrcell[asr]{4} & \asrcell[asr]{14.7}[1.8] & \asrcell[asr]{20} & \asrcell[asr]{14.0}[1.2] & \asrcell[asr]{26} & \asrcell[asr]{2} & 1/50 \\
\midrule
CACTUS & \asrcell[asr]{6.7}[0.7] & \asrcell[asr]{8} & \asrcell[asr]{17.3}[0.7] & \asrcell[asr]{24} & \asrcell[asr]{11.3}[2.4] & \asrcell[asr]{16} & \asrcell[asr]{60} & 30/50 \\
ChemToolAgent & \asrcell[asr]{4.7}[0.7] & \asrcell[asr]{8} & \asrcell[asr]{18.0}[1.2] & \asrcell[asr]{28} & \asrcell[asr]{20.0}[2.3] & \asrcell[asr]{32} & \asrcell[asr]{60} & 30/50 \\
Biomni & \asrcell[asr]{10.7}[1.8] & \asrcell[asr]{14} & \asrcell[asr]{22.0}[1.2] & \asrcell[asr]{34} & \asrcell[asr]{29.3}[1.8] & \asrcell[asr]{40} & \asrcell[asr]{56} & 28/50 \\
STELLA & \asrcell[asr]{4.0}[0.0] & \asrcell[asr]{4} & \asrcell[asr]{13.3}[1.3] & \asrcell[asr]{18} & \asrcell[asr]{16.7}[2.9] & \asrcell[asr]{26} & \asrcell[asr]{54} & 27/50 \\
DrugAgent & \asrcell[asr]{40.0}[3.1] & \asrcell[asr]{50} & \asrcell[asr]{40.7}[2.4] & \asrcell[asr]{52} & \asrcell[asr]{38.7}[2.7] & \asrcell[asr]{50} & \asrcell[asr]{38} & 19/50 \\
\bottomrule
\end{tabular}%
}
\end{table*}
\begin{table*}[!htbp]
\centering
\small
\setlength{\tabcolsep}{3pt}
\renewcommand{\arraystretch}{1.15}
\caption{Absolute ASR (\%) for GPT-5.6-Luna. Single-turn columns give mean $\pm$ standard error (SE) across three runs and pass@3; PAIR gives ASR and successful requests $k/50$ from one adaptive evaluation. PAIR ASR ranges from 28--66\% across harnesses, compared with 10\% for the base model. Green indicates lower ASR; orange and red indicate higher ASR.}
\label{tab:asr-gpt56l}
\resizebox{0.9\textwidth}{!}{%
\begin{tabular}{@{}l cc cc cc cc@{}}
\toprule
& \multicolumn{2}{c}{\textbf{Direct}}
& \multicolumn{2}{c}{\textbf{Roleplay}}
& \multicolumn{2}{c}{\textbf{Past tense}}
& \multicolumn{2}{c}{\textbf{PAIR}} \\
\cmidrule(lr){2-3}\cmidrule(lr){4-5}\cmidrule(lr){6-7}\cmidrule(lr){8-9}
\textbf{GPT-5.6-Luna}
& Mean $\pm$ SE & pass@3
& Mean $\pm$ SE & pass@3
& Mean $\pm$ SE & pass@3
& ASR & $k/50$ \\
\midrule
Base model & \asrcell[asr]{6.0}[2.0] & \asrcell[asr]{10} & \asrcell[asr]{12.0}[2.3] & \asrcell[asr]{18} & \asrcell[asr]{26.7}[0.7] & \asrcell[asr]{36} & \asrcell[asr]{10} & 5/50 \\
\midrule
CACTUS & \asrcell[asr]{9.3}[0.7] & \asrcell[asr]{14} & \asrcell[asr]{12.0}[2.3] & \asrcell[asr]{18} & \asrcell[asr]{16.7}[3.3] & \asrcell[asr]{26} & \asrcell[asr]{66} & 33/50 \\
ChemToolAgent & \asrcell[asr]{4.7}[1.8] & \asrcell[asr]{10} & \asrcell[asr]{10.7}[4.8] & \asrcell[asr]{20} & \asrcell[asr]{8.7}[3.7] & \asrcell[asr]{20} & \asrcell[asr]{42} & 21/50 \\
Biomni & \asrcell[asr]{5.3}[0.7] & \asrcell[asr]{8} & \asrcell[asr]{14.0}[5.0] & \asrcell[asr]{24} & \asrcell[asr]{22.7}[0.7] & \asrcell[asr]{36} & \asrcell[asr]{50} & 25/50 \\
STELLA & \asrcell[asr]{4.7}[0.7] & \asrcell[asr]{8} & \asrcell[asr]{8.0}[3.1] & \asrcell[asr]{16} & \asrcell[asr]{22.7}[4.1] & \asrcell[asr]{40} & \asrcell[asr]{50} & 25/50 \\
DrugAgent & \asrcell[asr]{14.7}[1.3] & \asrcell[asr]{28} & \asrcell[asr]{18.7}[2.4] & \asrcell[asr]{28} & \asrcell[asr]{19.3}[1.3] & \asrcell[asr]{36} & \asrcell[asr]{28} & 14/50 \\
\bottomrule
\end{tabular}%
}
\end{table*}
\begin{table*}[!htbp]
\centering
\small
\setlength{\tabcolsep}{3pt}
\renewcommand{\arraystretch}{1.15}
\caption{Absolute ASR (\%) for GPT-6-Luna. Single-turn columns give mean $\pm$ standard error (SE) across three runs and pass@3; PAIR gives ASR and successful requests $k/50$ from one adaptive evaluation. PAIR ASR ranges from 8--30\% across harnesses, compared with 2\% for the base model. Green indicates lower ASR; orange and red indicate higher ASR.}
\label{tab:asr-gpt6l}
\resizebox{0.9\textwidth}{!}{%
\begin{tabular}{@{}l cc cc cc cc@{}}
\toprule
& \multicolumn{2}{c}{\textbf{Direct}}
& \multicolumn{2}{c}{\textbf{Roleplay}}
& \multicolumn{2}{c}{\textbf{Past tense}}
& \multicolumn{2}{c}{\textbf{PAIR}} \\
\cmidrule(lr){2-3}\cmidrule(lr){4-5}\cmidrule(lr){6-7}\cmidrule(lr){8-9}
\textbf{GPT-6-Luna}
& Mean $\pm$ SE & pass@3
& Mean $\pm$ SE & pass@3
& Mean $\pm$ SE & pass@3
& ASR & $k/50$ \\
\midrule
Base model & \asrcell[asr]{2.7}[0.7] & \asrcell[asr]{4} & \asrcell[asr]{4.0}[0.0] & \asrcell[asr]{4} & \asrcell[asr]{8.7}[2.4] & \asrcell[asr]{16} & \asrcell[asr]{2} & 1/50 \\
\midrule
CACTUS & \asrcell[asr]{2.0}[0.0] & \asrcell[asr]{2} & \asrcell[asr]{4.7}[1.8] & \asrcell[asr]{8} & \asrcell[asr]{6.7}[1.3] & \asrcell[asr]{10} & \asrcell[asr]{24} & 12/50 \\
ChemToolAgent & \asrcell[asr]{2.7}[0.7] & \asrcell[asr]{4} & \asrcell[asr]{3.3}[1.8] & \asrcell[asr]{8} & \asrcell[asr]{6.7}[1.3] & \asrcell[asr]{14} & \asrcell[asr]{26} & 13/50 \\
Biomni & \asrcell[asr]{2.0}[1.2] & \asrcell[asr]{4} & \asrcell[asr]{4.7}[1.8] & \asrcell[asr]{8} & \asrcell[asr]{4.0}[2.0] & \asrcell[asr]{8} & \asrcell[asr]{28} & 14/50 \\
STELLA & \asrcell[asr]{2.0}[0.0] & \asrcell[asr]{2} & \asrcell[asr]{4.0}[1.2] & \asrcell[asr]{10} & \asrcell[asr]{10.7}[2.4] & \asrcell[asr]{22} & \asrcell[asr]{30} & 15/50 \\
DrugAgent & \asrcell[asr]{1.3}[0.7] & \asrcell[asr]{4} & \asrcell[asr]{3.3}[1.3] & \asrcell[asr]{6} & \asrcell[asr]{8.0}[2.3] & \asrcell[asr]{18} & \asrcell[asr]{8} & 4/50 \\
\bottomrule
\end{tabular}%
}
\end{table*}

\subsection{Harness Configuration Ablations}
\label{app:ablations}

The main-text comparisons measure the effect of introducing a
complete harness. Here, we explore whether disabling selected
components changes harmful compliance in STELLA and Biomni.
We focus on PAIR with GPT-5.5 and GPT-6-Luna, where both
complete harnesses substantially increase ASR relative to the
base models. These comparisons concern modified configurations
and do not isolate individual component effects.

\begin{table}[!h]
\centering
\small
\setlength{\tabcolsep}{6pt}
\renewcommand{\arraystretch}{1.15}

\caption{
PAIR ASR (\%) for complete and modified harness configurations,
with $\Delta$ASR relative to the corresponding base LLM
(percentage points). Each configuration is evaluated on 50 requests.
STELLA retains its code interpreter when external tools are
disabled. Absolute ASR is colored from green (low) to red (high),
while positive $\Delta$ASR uses the orange/red scale of the main table.
}
\label{tab:exploratory-configurations}

\begin{tabular}{@{}ll cc cc@{}}
\toprule
& &
\multicolumn{2}{c}{\textbf{GPT-5.5}} &
\multicolumn{2}{c}{\textbf{GPT-6-Luna}} \\
\cmidrule(lr){3-4}
\cmidrule(l){5-6}

\textbf{System} & \textbf{Configuration}
& ASR & $\Delta$ASR
& ASR & $\Delta$ASR \\
\midrule

Base LLM & No harness
& \asrcell[asr]{2} & --
& \asrcell[asr]{2} & -- \\

\midrule

STELLA & Full harness
& \asrcell[asr]{54} & \asrcell{+52}
& \asrcell[asr]{30} & \asrcell{+28} \\

& External tools disabled
& \asrcell[asr]{52} & \asrcell{+50}
& \asrcell[asr]{28} & \asrcell{+26} \\

& Memory disabled
& \asrcell[asr]{56} & \asrcell{+54}
& \asrcell[asr]{34} & \asrcell{+32} \\

\midrule

Biomni & Full harness
& \asrcell[asr]{56} & \asrcell{+54}
& \asrcell[asr]{28} & \asrcell{+26} \\

& Tools disabled
& \asrcell[asr]{60} & \asrcell{+58}
& \asrcell[asr]{22} & \asrcell{+20} \\

\bottomrule
\end{tabular}
\end{table}

Disabling these components produces relatively small changes
compared with the difference between each complete harness and
its base model. With GPT-5.5, STELLA increases PAIR ASR from
2\% for the base model to 54\%. Disabling external tools
reduces ASR to 52\%, while disabling memory raises it to 56\%.
Biomni reaches 56\% with the full harness and 60\% without
tools. With GPT-6-Luna, disabling Biomni tools reduces ASR
from 28\% to 22\%, still well above the 2\% baseline.
The changes are not consistent across models or components.

These results also illustrate why conventional component
ablations are difficult for agent harnesses. Tools, memory,
system instructions, and orchestration are closely connected.
STELLA, for example, retains its code interpreter when
external tools are disabled because the interpreter is also
used to produce its final response. Removing an interface
can disrupt the workflow, while changing instructions or
routing to accommodate its removal also changes the system
being evaluated.

We therefore interpret these results as exploratory
configuration comparisons. They show that substantial
safety regressions remain in the modified systems, but
do not identify which components cause them. More controlled
experiments would be needed to separate those effects and
determine how harnesses can preserve base-model safety.

\end{document}